\documentclass[runningheads]{llncs}
\usepackage{amsmath}
\usepackage{amssymb}
\usepackage{graphicx}
\usepackage{enumitem}
\usepackage{color}

\usepackage{hyperref}
\usepackage{url}

\begin{document}
\title{Jointly Learning to Detect Emotions and Predict Facebook Reactions 
	\thanks{This is a post-peer-review, pre-copyedit version of an article published in LNCS, volume 11730. The final authenticated version is available online at:
		\url{https://doi.org/10.1007/978-3-030-30490-4_16}
	}
	}

\author{Lisa Graziani[0000-0002-7384-9633]* \inst{1} \and
Stefano Melacci[0000-0002-0415-0888] \inst{2} \and
Marco Gori[0000-0001-6337-5430] \inst{2}}

\authorrunning{Lisa Graziani, Stefano Melacci, Marco Gori}

\institute{DINFO, University of Florence, Italy,
	\email{lisa.graziani@unifi.it} \and
DIISM, University of Siena, Italy, \email{\{mela,marco\}@diism.unisi.it}}

\maketitle              

\begin{abstract}
The growing ubiquity of Social Media data offers an attractive perspective for improving the quality of machine learning-based models in several fields, ranging from Computer Vision to Natural Language Processing. In this paper we focus on Facebook posts paired with ``reactions'' of multiple users, and we investigate their relationships with classes of emotions that are typically considered in the task of emotion detection. We are inspired by the idea of introducing a connection between reactions and emotions by means of First-Order Logic formulas, and we propose an end-to-end neural model that is able to jointly learn to detect emotions and predict Facebook reactions in a multi-task environment, where the logic formulas are converted into polynomial constraints. Our model is trained using a large collection of unsupervised texts together with data labeled with emotion classes and Facebook posts that include reactions. An extended experimental analysis that leverages a large collection of Facebook posts shows that the tasks of emotion classification and reaction prediction can both benefit from their interaction.

\end{abstract}

\section{Introduction}
\label{sec:intro}

Social media have strongly changed the way we interact with each other and how we share contents.
Many people exploit social networks to publish details of their daily lives, their opinions and their thoughts. These data represent a precious source of information for building large datasets of annotated multimedia contents, or for mining users' behaviours and other user-related information.

A valuable feature for every modern system that interacts with humans is understanding the emotional state of users. Conversational systems can adapt their language in function of the perceived user emotions, digital marketing platforms can customize recommendations, social media marketing strategies can be changed in function of the estimated emotions triggered when posting contents. 
If we restrict our attention to the case of text, emotion detection is a widely studied and still challenging task \cite{strapparava2008learning,kim2010evaluation,agrawal2012unsupervised,mohammad2012portable,herzig2017emotion}. 
In the case of categorical emotion detection, sentences are usually classified into the six universal emotions defined by Ekman \cite{ekman1971constants}, namely \emph{anger, disgust, fear, happiness, sadness}, and \emph{surprise}.

This paper is rooted on the connections between the task of emotion detection and social media data. There is an intrinsic link between certain categories of tags attached to user posts and the emotional state of those users that participate in the tagging process. We focus on the case of Facebook, where users can express their feeling on a post through the so called ``reactions'', that are \texttt{LOVE, HAHA, WOW, SAD, ANGRY}, together with the widely known \texttt{LIKE}. While \texttt{LIKE} represents a universal and generic expression of a positive feedback, the other reactions are more fine-grained, and somewhat related to the aforementioned categories of emotions. However, this relationship is weak and distant, since some reactions can be loosely associated to emotional categories, sometimes with large ambiguity. For example, \texttt{WOW} expresses ``surprise'' but it can be also used to describe contents where the astonishment is accompanied by ``fear''. Moreover, Facebook reactions are the outcome of a tagging process where users might follow superficial and strongly subjective criteria to react.

Recently, Facebook reactions have been studied in the context of emotion detection. Some authors trained emotion classification models using Facebook reactions \cite{raad2018aseds}, \cite{pool2016distant}, while others tried to learn to predict Facebook reactions in a given domain, bootstrapping the system with the outcome of   emotion mining \cite{krebs2017social}. Reactions are usually manually mapped to (a subset of) the aforementioned universal emotions, providing a form of distant supervision. Differently, the task of emotion detection from text has been the subject of a large number of studies, mostly distinguished into lexicon-based and machine learning-based approaches (or hybrid solutions). Lexicon-based approaches employ linguistic models or prior knowledge for the classification task, and they essentially give a score to a sentence using a predefined sentiment lexicon, without using labeled data \cite{kim2010evaluation}, \cite{strapparava2008learning}. In \cite{agrawal2012unsupervised} the authors propose an unsupervised context-based emotion detection method that does not rely on any affect dictionaries or annotated training data.
A constraint optimization framework based on lexicon is presented in \cite{wang2015detecting}.
Machine learning-based methods usually exploit supervised learning algorithms trained on annotated corpora. 
The approach of \cite{qadir2014learning} focusses on Twitter data, while \cite{chaffar2011using} uses a heterogeneous emotion-annotated dataset to recognize the six basic emotions.
Finally, \cite{herzig2017emotion} focusses on an ensemble model, strongly exploiting pre-trained, dense word-embedding representations.
 
In this paper we propose a neural network-based model to jointly learn the task of emotion detection and the task of predicting Facebook reactions. Our model consists of a bidirectional Long Short-Term Memory (LSTM) recurrent neural network \cite{schuster1997bidirectional,hochreiter1997long} to encode the input sentence, and two predictors associated with the considered tasks. Predictors are not independent, but are linked by prior knowledge on the relationships between the tasks. Such knowledge is represented by First-Order Logic (FOL) formulas, which allow us to naturally express how reactions are connected to emotion classes and vice-versa. 
Following the framework of Learning from Constraints \cite{gnecco2015foundations}, FOL formulas are converted into polynomial constraints and softly enforced into the learning problem, thus tolerating some violations.
The system automatically learns ``how'' to fulfil the FOL formulas in function of the way the data are distributed.
Our model is trained using a heterogeneous dataset composed of data labeled with emotion classes, Facebook posts that include user reactions, and a large collection of unsupervised posts. We do not use any external lexical resources, and an extended experimental analysis shows that the tasks of emotion classification and reaction prediction can both benefit from their interaction. The resulting emotion detector is competitive with some models that exploit lexical resources or ad-hoc features, and we also investigate the role of pre-trained word embeddings.

This paper is organized as follows. Section~\ref{sec:model} describes the proposed model, while Section \ref{sec:logic} focusses on the logic constraints. Experimental results are provided in Section~\ref{sec:results} and, finally, Section~\ref{sec:conclusions} concludes the paper with our comments.

\section{Model and Data Organization}
\label{sec:model}

We consider a multi-task setting where two predictors $p_{r}(x)$ and $p_{e}(x)$ operate on the same data $x$, that is a short input text. Such predictors are associated to the task of reaction classification ($p_{r}$) and emotion classification ($p_{e}$), respectively. In the context of this paper, both the tasks consist in predicting the most dominant reaction/emotion when processing a text $x$.\footnote{Some approaches consider these tasks as multi-label prediction problems \cite{kim2010evaluation,strapparava2008learning}, while other authors focus on the most dominant response \cite{chaffar2011using}, as we do in this paper. What we propose can be adapted to the case of multi-label prediction.}
In detail, $p_{r}(x) \in [0,1]^{R}$ outputs a probability distribution over $R$ reactions, and, analogously, $p_{e}(x) \in [0,1]^{E}$ outputs a probability distribution over $E$ classes of emotions. We select the emotion-reaction pair associated to the largest probabilities.

\begin{figure}
	\centering
	\includegraphics[width=0.6\textwidth]{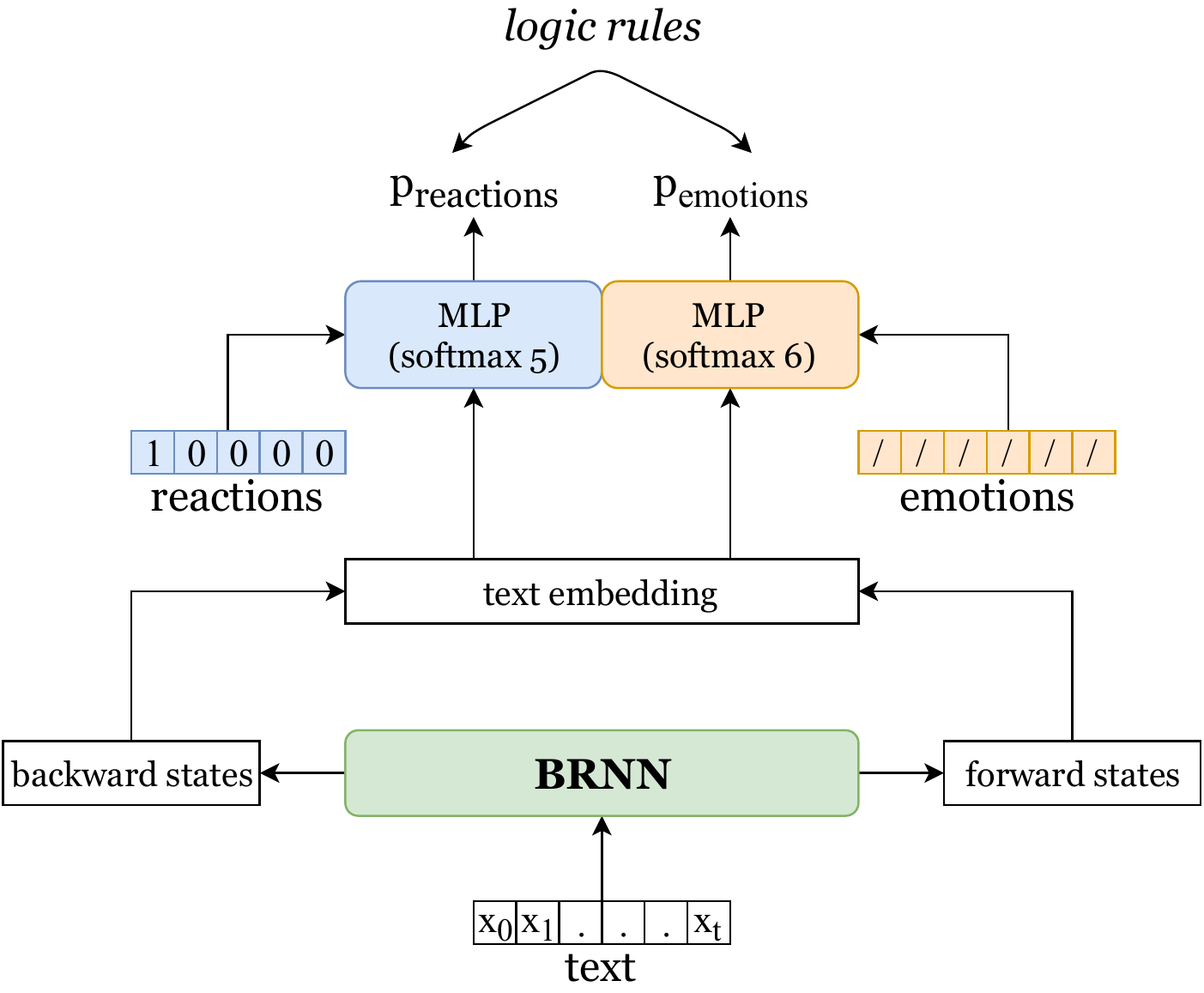}
	\caption{The proposed model. Predictors $p_r$ and $p_e$ are reported with their expanded names: $p_{reactions}$ and $p_{emotions}$. When training the network, we feed it with text either labeled with emotions or reactions, and logic constraints bridge the two predictors.}
	\label{model_structure}
\end{figure}

Following the classical pipeline of several machine learning-based approaches in Natural Language Processing, the input text $x$ is tokenized into words $x_0,\ldots,x_t$ belonging to a fixed-size vocabulary. Each word is embedded into a learnable latent dense representation, also known as ``word embedding'', and a Long Short-Term Memory (LSTM) recurrent neural network \cite{hochreiter1997long} processes the sequence of word embeddings in both directions (Bidirectional Recurrent Neural Network (BRNN) \cite{schuster1997bidirectional}). The forward and backward states are then concatenated, producing an embedded latent representation of $x$, that is provided as input to two Multi Layer Perceptrons (MLPs) with softmax activation functions in the output layers, thus implementing $p_{r}$ and $p_{e}$, respectively. The choice of sharing the same latent representation of $x$ with both predictors is due to the fact that the two prediction tasks are  certainly correlated.
Finally, during the training stage, the MLPs are connected by constraints that are devised from FOL rules, and that will be described in Section~\ref{sec:logic}.
The whole architecture is reported in Fig. \ref{model_structure}.

\begin{figure}
	\centering
	\includegraphics[width=0.6\textwidth]{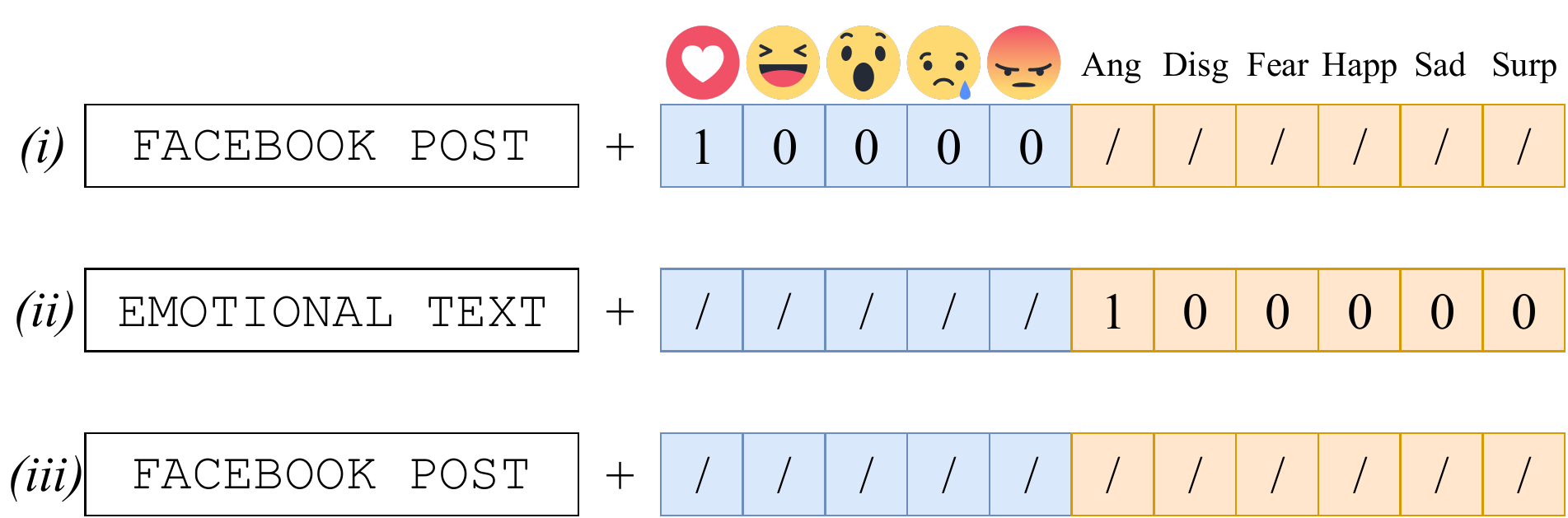}
	\caption{Sample representatives of the types of data included in our heterogeneous training set. $(i)$ A facebook post paired with the reaction label \texttt{LOVE} (encoded with the blue 1-hot vector) and no emotion labels. $(ii)$ Text paired with the emotion class \textit{anger} (orange 1-hot vector) and no reaction labels. $(iii)$ An unlabeled Facebook post. } \label{data_img}
\end{figure}

Our model is trained using a heterogeneous collection of text $\mathcal{T}$ of partially labeled and unlabeled data, composed by the union of three disjoint sets, $\mathcal{T}_r,\ \mathcal{T}_e,\ \mathcal{T}_u$, that, in turn, consist of pairs $(x,y)$, where $y$ is either a reaction label, an emotion label, or a dummy placeholder (i.e., unlabeled data), respectively. $(i)$ The set $\mathcal{T}_r$ is a collection of Facebook posts, each of them labeled with one out of $R=5$ reaction classes (listed in Section~\ref{sec:intro}), encoded with a one-hot vector $y_r$ of size $R$.  
We did not consider the class \texttt{LIKE}, since it is too generic, and we selected the most frequent reaction class in each post. Moreover, $\mathcal{T}_r$ is composed only by those posts with at least $\tau$ reaction hits in total ($\tau=20$ in our experience), and where the most frequent reaction has a number of hits that is  greater than the number of hits of all the other reactions scaled by a factor $\gamma$ (we set $\gamma=0.4$).
$(ii)$ The set $\mathcal{T}_e$ is a collection of sentences, each of them labeled with one of the $E=6$ universal emotions (see Section~\ref{sec:intro}), encoded with a one-hot vector $y_e$ of size $E$.\footnote{In our experience we did not consider the \textit{neutral} class, that, however, could be easily introduced in the proposed model.} We exploited existing databases to build $\mathcal{T}_e$ (see Section~\ref{sec:results}), keeping only the most dominant emotion in the case of multi-labeled data.
$(iii)$ Finally, the set $\mathcal{T}_u$ is a collection of unlabeled text,
 that in our experience, consists of a large collection of Facebook posts without reactions. Each sample is paired with a dummy label vector $y_{\texttt{none}}$.
This set is exploited to enforce the logic constraints (Section~\ref{sec:logic}) in space regions that are not covered by the labeled portion(s) of the training set. This allows the model to learn predictors that better generalize the information associated to the logic formulas.
A sketch that summarizes the types of training data used in this paper is reported in Fig.~\ref{data_img}.

\section{Multi-Task Learning with Constraints}
\label{sec:logic}
Before introducing the approach that we propose with this paper, we mention that the simplest way to bridge the tasks of emotion and reaction classification is to generate artificial labels, i.e., to define a fixed mapping between emotions and reactions and augment the training data with these new labels (see, for example \cite{pool2016distant}, Table 1). Considering the emotion/reaction classes of Section~\ref{sec:model}, a reasonable mapping from reactions to emotions, represented with the notation ``ground truth'' $\to$ ``new label'', is the following one: \texttt{LOVE} $\to$ \emph{happiness}, \texttt{WOW} $\to$ \emph{surprise}, \texttt{HAHA} $\to$ \emph{happiness}, \texttt{SAD} $\to$ \emph{sadness}, \texttt{ANGRY} $\to$ \emph{anger}. Similarly, we can map emotions to reactions: \emph{anger} $\to$ \texttt{ANGRY}, \emph{disgust} $\to$ \texttt{ANGRY}, \emph{fear} $\to$ \texttt{WOW}, \emph{happiness} $\to$ \texttt{HAHA}, \emph{sadness} $\to$ \texttt{SAD}, \emph{surprise} $\to$ \texttt{WOW}.
However, this manual conversion is rigid and sometimes ambiguous. For example, no reactions are converted into labels of classes \emph{fear} and \emph{disgust}, and no emotions are mapped into the reaction \texttt{LOVE}.

We propose to describe the mappings between emotion and reaction classes using FOL formulas and to develop a multi-task system that learns from them, following the framework of Learning from Constraints  \cite{gnecco2015foundations,gori2013constraint,graziani2018role}. Each class is associated to a predicate, whose truth degree is computed using a function that, for simplicity, we indicate with the name of the class itself. These predicates can be seen as the components of the vectorial functions $p_r(x)$ and $p_e(x)$, i.e., $p_r(x) = \left[\texttt{HAHA}(x),\texttt{SAD}(x), \texttt{ANGRY}(x), \texttt{LOVE}(x), \texttt{WOW}(x) \right]$, and $p_e(x) = \left[\emph{anger}(x),\emph{disgust}(x),\right.$ $\left.\emph{fear}(x),\emph{happiness}(x),\emph{sadness}(x),\emph{surprise}(x) \right]$. We define the following rules,
\begin{eqnarray}
	\label{uno}\forall x \, \texttt{HAHA}(x)& \Rightarrow&\emph{happiness}(x) \\
	[-1mm]\label{due}\forall x \, \texttt{SAD}(x)& \Rightarrow& {sadness}(x)\\ 
	[-1mm]\label{tre}\forall x \,\texttt{ANGRY}(x)& \Rightarrow&  \emph{anger}(x) \; \lor \; \emph{disgust}(x)\\ 
	[-1mm]\label{quattro}\forall x \,\texttt{LOVE}(x)& \Rightarrow& \emph{happiness}(x)\\ 
	[-1mm]\label{cinque}\forall x \,\texttt{WOW}(x)& \Rightarrow& \emph{surprise}(x) \; \lor \; \emph{fear}(x)\\ 
	[-1mm]\label{set}\forall x \,\emph{anger}(x)& \Rightarrow& \texttt{ANGRY}(x)\\ 
	[-1mm]\label{sette}\forall x \,\emph{disgust}(x)& \Rightarrow& \texttt{ANGRY}(x)\\ 
	[-1mm]\label{otto}\forall x \,\emph{fear}(x)& \Rightarrow& \texttt{WOW}(x)\\ 
	[-1mm]\label{nove}\forall x \,\emph{happiness}(x)& \Rightarrow& \texttt{HAHA}(x) \; \lor \; \texttt{LOVE}(x)\\ 
	[-1mm]\label{dieci}\forall x \,\emph{sadness}(x)& \Rightarrow& \texttt{SAD}(x)\\ 
	[-1mm]\label{undici}\forall x \,\emph{surprise}(x)& \Rightarrow& \texttt{WOW}(x) \ .
\end{eqnarray}
Notice that these rules do not include negations, that is due to the probabilistic relationship (softmax) that we introduced in the output of the predictors (if a function goes toward $1$, all the others will automatically go toward $0$).\footnote{We did not write the rules in a more compact form using the double implication $\Leftrightarrow$, since we will differently weigh the impact of some of them, as it will be clear shortly.}

We defined our FOL formulas after having analyzed the content of various Facebook posts and the associated reactions. Implications \ref{tre}-\ref{cinque}-\ref{nove} include an ambiguous mapping, modeled with the $\lor$ operator (disjunction). The second predicate that we reported in each disjunction corresponds to a less trivial mapping that, at a first glance, might not always seem obvious. However, in our experience, we found these cases to be more frequent than expected. We report an example for each of them:
\texttt{WOW} could be \emph{fear} instead of \emph{surprise} (Eq. \ref{cinque}),
\begin{quote}
\footnotesize
	\emph{Snake on a plane: Frightening moment on an Aeromexico flight when a large snake fell from overhead mid-flight. The flight made a quick landing and animal control took the stowaway into custody.} 
\end{quote}
Emotion \emph{happiness} could be converted into \texttt{LOVE} instead of \texttt{HAHA} (Eq. \ref{nove}), 
\begin{quote}
\footnotesize
\emph{When I got a wedding ring of diamond from the boy I loved.}
\end{quote}
The reaction \texttt{ANGRY} could be eventually mapped into \emph{disgust} (Eq. \ref{tre}),
\begin{quote}
\footnotesize
	\emph{The San Antonio police chief said that former officer Matthew Luckhurst committed a vile and disgusting act that violates our guiding principles.}
\end{quote}

Our rules are converted into real-valued polynomials by means of T-Norms \cite{gnecco2015foundations,gori2013constraint}, that are functions modeling the logical AND whose output is in $ [0,1]$. We used the Product T-Norm, where the logical AND is simply the product of the involved arguments. In turn, this choice transforms $a \Rightarrow b$ into the polynomial $1-a+a\cdot b$ (see \cite{gnecco2015foundations,gori2013constraint} for further details).
Constraining the FOL formula to hold true leads to enforcing the T-Norm-based polynomials to be $1$, so we get equality constraints, e.g., $1-a+a\cdot b = 1$ in the previous example.
We introduce these constraints into the learning problem in a soft manner using penalty functions, so that the system might decide to violate some of them for some input $x$ (in our implementation, we used the penalty $-\log(\cdot)$). 

Formally, the multi-task function that we minimize to learn the model is 
\begin{equation}\label{loss}
 \sum_{(x,y_r) \in \mathcal{T}_{r}}  \hskip-3mm  L\left(p_r(x),y_r\right) +  \hskip-3mm  \sum_{(x,y_e) \in \mathcal{T}_{e}} \hskip-3mm  L \left(p_e(x),y_{e}\right) + \sum_{j=1}^{11}\sum_{(x,y_{\texttt{none}}) \in \mathcal{T}}  \hskip-3mm   w_j \phi_j\left(p_r(x),p_e(x) \right) \ ,
\end{equation}
where we avoided reporting the scaling factors in front of each term of the summation, to keep the notation simpler. The function $L$ is the cross-entropy loss, while $\phi_j$ is the penalty term associated to the $j$-th FOL formula, weighed by the scalar $w_j > 0$.\footnote{Each $\phi_j$ might only consider some of the output components of $p_r(x)$ and $p_e(x)$, depending on the FOL formula that it implements.} Notice that FOL formulas are constrained to hold true on all the available training data, including the large collection of unlabeled text $\mathcal{T}_u$. This allows the system to learn predictors that fulfil the FOL rules in regions of the input space that might not be covered by the labeled data, thus increasing the information transfer between the two tasks (as typically done in the framework of Learning from Constraints \cite{gnecco2015foundations,gori2013constraint}). 
Thanks to this formulation, we can differently weigh the impact of each constraint in function of the confidence we have on it, tuning the parameters $w_j$. For example, constraints associated to Formulas \ref{quattro}-\ref{sette}-\ref{otto} are weaker that the other ones, and we decided to keep their weight small.

\section{Experimental Results}
\label{sec:results}
In order to evaluate the proposed model, we created a heterogeneous data collection that follows the organization described in Section~\ref{sec:model}.
In particular, we considered a large public dataset of \emph{Facebook posts} that are scraped from Facebook pages of newspapers.
\footnote{\tiny\url{https://data.world/martinchek/2012-2016-facebook-posts}}
Data was filtered accordingly to what we described in Section~\ref{sec:model}, ending up with $\approx 200,000$ posts, out of which $100,000$ are left unlabeled.
Then, we collected the most popular datasets containing text labeled with emotions, namely 
 \emph{AffectiveText}, \emph{ISEAR}, and \emph{Fairy Tales}.  
\emph{AffectiveText} (SemEval-2007 \cite{strapparava2007semeval}) contains 1,250 short newspaper headlines.
Sentences are labeled with the six basic emotions, and each of them is scored in a range from 0 to 100.    
For the purpose of this experimentation, we took the emotion with the highest score.
\emph{ISEAR} (International Survey on Emotion Antecedents and Reactions \cite{scherer1994evidence})
contains 7,666 sentences from questionnaires about emotional experiences covering anger, disgust, fear, joy, sadness, shame, guilt. We discarded the last two classes since they are not part of the universal emotions, and mapped ``joy'' to ``happiness'' (the class ``surprise'' is missing).
\emph{Fairy Tales} \cite{alm2008affect} 
contains sentences belonging to short stories, annotated with multiple labels. We discarded the neutral class and we kept only sentences with four identical labels (three for the class ``disgust'', due to the small number of samples). In Table \ref{data_number} we report the details of the data exploited in this paper. 
\begin{table}
	\centering
	\caption{Number of Facebook posts for each reaction, and number of \textit{unlabeled} posts (top). Number of texts for each emotion class, covering three public datasets (bottom).}\label{data_number}
	\scalebox{0.92}{
	\begin{tabular}{|l|c c c c c c |c|}
		\hline
		& \texttt{LOVE} & \texttt{WOW} & \texttt{HAHA} & \texttt{SAD} & \texttt{ANGRY} & \textit{Unlabeled} &$\ $\textsc{Total}$\ $\\
				\hline
		{Facebook Posts} &  31801 & 13807 & 17552 & 16689 & 15775 & 100000 &  195624 \\
		\hline
		\hline
		& \emph{Anger} & \emph{Disgust} & \emph{Fear} & \emph{Happiness} & \emph{Sadness} & \emph{Surprise} & $\ $\textsc{Total} $\ $\\
				\hline
		{Affective Text} &  91 & 42 & 194 & 441 & 265 & 217 & 1250\\
		{ISEAR} & 1087 & 1082 & 1089 & 1090 & 1083 & 0 & 5431\\
		{Fairy Tales} & 146 & 64 & 166 & 445 & 264 & 100 & 1185\\
		\hline
	\end{tabular}}
\end{table}

We evenly divided our heterogeneous datasets into 3 splits, keeping the original data distribution among classes.
Each split is further divided into training, validation and test sets, with special attention in preparing the test data.
In particular, the test set is composed of 15\% of the labeled Facebook posts, merged with one of ISEAR, Fairy Tales, Affective Text. 
As a matter of fact, each of such emotional datasets is small sized (considering the number of classes and the intrinsic difficulty of the learning task), and it has different properties w.r.t. the other two ones. We experienced that training and testing on subportions of the same emotional dataset leads to performances that do not reflect the concrete quality of the system when it is deployed and tested in a generic context. Differently, training and testing on different emotional datasets offers a more realistic perspective of the generalization quality of the resulting system. The training set includes 70\% of the labeled Facebook posts and 80\% of the two emotional datasets which are not present in the test set, plus the unlabeled Facebook posts. The validation set is composed of the remaining data, that is, 15\% of labeled posts and 20\% of the two emotional datasets which are not used as test set.
We preprocessed all the data converting text to lowercase, removing URLs, standardizing numbers with a special token, removing brackets, 
 separating punctuation and hashtags.
Then, we created a vocabulary composed of the most frequent $10,000$ words and we truncated sentences longer than 30 words, to make them more easily manageable by the BRNN.

We evaluated architectures with differently sized word embeddings (from 50 to 300 units each), states of the BRNN (in the range $[50,200]$), hidden layers (and number of units) of the final MLPs (up to $2$ hidden layers). 
After a first exploratory experimentation, we focussed on models with word embeddings of size 100, BRNN with a hidden state composed of 100 units and final predictors with no hidden layers, that were providing the best results in the validation data. Then, we kept validating in more detail all the other model parameters (learning rate, the possibility of introducing drop-out right after the BRNN, weight of the logic constraints $w_j$). We considered the (macro) F1 scores on each task to evaluate the quality of our models, and we early stopped the training procedure whenever the average F1 score on the validation data was not increased after $20$ epochs (keeping the model associated to the best F1 score found so far).

We compared the following models:
\begin{itemize}[leftmargin=*,labelindent=4.5em]
\item[\footnotesize \textsc{Plain}.] The model of Fig.~\ref{model_structure}, without logic constraints ($w_j = 0$, $\forall j$).
\item[\footnotesize \textsc{Constr}.] The same as \textsc{Plain}, but including logic constraints ($w_j > 0$, $\forall j$).
\item[\footnotesize \textsc{Artificial}.] The same as \textsc{Plain}, where the training data is augmented with artificially mapped classes as described at the beginning of Section~\ref{sec:logic}.
\item[\footnotesize \textit{+Emb}.] Variant of the models above, based on pre-trained word embeddings of size $300$ (the popular Google word2vec model).\footnote{In this case, after our initial exploratory experimentation, we selected a BRNN with state size $200$, and reaction predictor with a hidden layer of size $25$.}
\end{itemize}

We first evaluate the quality of the system in the task of reaction prediction. In Table \ref{reaction_table}, we can appreciate how introducing logic constraints constantly improves the quality of the predictor in all the reaction classes. Using artificial labels from emotional data is far from giving the same benefits of logic constraints, and we did not experience advantages in using pre-trained word embeddings, that might be due to the inherent noise in the reaction prediction task.
\begin{table}
	\centering
	\caption{F1 scores on Facebook reactions (test data, averaged over the 3 data splits - std dev. in bracket). Bold: cases in which constraints introduce improvements.}\label{reaction_table}
	\scalebox{0.92}{
	\begin{tabular}{|l|c c c c c |c|}
		\cline{2-7}
		\multicolumn{1}{c|}{}&  \texttt{LOVE} & \texttt{WOW} & \texttt{HAHA} & \texttt{SAD} & \texttt{ANGRY}  & Macro Avg \\
		\cline{2-7}
		\hline
		\textsc{Plain} & 0.630 \tiny{(0.009)} & 0.354 \tiny{(0.008)} & 0.440 \tiny{(0.009)} & 0.532 \tiny{(0.014)} & 0.329 \tiny{(0.012)} & 0.457 \tiny{(0.007)}  \\
		\textsc{Constr} &  \textbf{0.639} \tiny{(0.162)} & \textbf{0.371} \tiny{(0.013)} & \textbf{0.443} \tiny{(0.003)} & \textbf{0.535} \tiny{(0.005)} & \textbf{0.347} \tiny{(0.007)} & \textbf{0.467} \tiny{(0.007)}\\
		 \textsc{Artificial} & 0.596 \tiny{(0.051)} & 0.324 \tiny{(0.015)} & 0.393 \tiny{(0.028)}  & 0.451 \tiny{(0.077)} & 0.303 \tiny{(0.030)} & 0.413 \tiny{(0.038)}\\ 
		\hline
		 \textsc{Plain}\emph{+Emb} &  0.614 \tiny{(0.019)} & 0.343 \tiny{(0.014)} & 0.425 \tiny{(0.012)} & 0.531 \tiny{(0.007)} & 0.345 \tiny{(0.013)} & 0.452 \tiny{(0.006)} \\
		\textsc{Constr}\emph{+Emb}  $\ $ &  \textbf{0.638} \tiny{(0.007)} & \textbf{0.347} \tiny{(0.003)} & \textbf{0.437} \tiny{(0.005)} & \textbf{0.538} \tiny{(0.012)} & \textbf{0.356} \tiny{(0.009)} & \textbf{0.463} \tiny{(0.003)}\\
		\textsc{Artif.}\emph{+Emb}  & 0.608 \tiny{(0.031)} & 0.323 \tiny{(0.006)} & 0.375 \tiny{(0.031)} & 0.446 \tiny{(0.070)} &  0.311 \tiny{(0.002)} & 0.412 \tiny{(0.030)} \\
		\hline
	\end{tabular}}
\end{table} 

Moving to the task of emotion classification, we report the results we obtained in the previously described test sets, that correspond to three different emotional datasets.
In Table \ref{isear_table} we focus on testing in the ISEAR data. Logical rules always allow the model to improve the macro-averaged F1 scores. We notice that the F1 score on ``disgust'' and ``fear'' classes is largely better than when not using constraints. In fact, without exploiting the logical rules of Eq. \ref{tre} and \ref{cinque} there is no transfer of information from reaction data, and the supervised portion of the training set is not enough to learn good predictors.
Interestingly, this consideration does not hold when using pre-trained embeddings, where the performances of the not-constrained model are already close to the constrained one. In this case, all the other classes are improved instead. Finally, artificial labels do not seem a promising solution. 
\begin{table}
	\centering
	\caption{F1 scores on emotion classification (ISEAR). Bold: cases in which constraints introduce improvements.}\label{isear_table}
	\scalebox{0.92}{
	\begin{tabular}{|l|c c c c c |c|}
		\cline{2-7}
		\multicolumn{1}{c|}{}&  \emph{Anger} & \emph{Disgust} & \emph{Fear} & \emph{Happiness} & \emph{Sadness}  & Macro Avg \\
		\cline{2-7}
		\hline
		\textsc{Plain}   & {0.313} & 0.009 & 0.170 & {0.452} & {0.420} & 0.227\\
		\textsc{Constr} & 0.200 & \textbf{0.185} & \textbf{0.272} & 0.395 & 0.419 &  \textbf{0.245}\\
		\textsc{Artificial} & 0.186 & 0.025 & 0.039 & 0.126 & 0.246 & 0.104 \\
		\hline
		\textsc{Plain}\emph{+Emb}   & 0.366 & {0.149} & {0.383} & 0.522 & 0.466 &  0.314\\
		\textsc{Constr}\emph{+Emb} $\ $ & \textbf{0.383} & 0.146 & 0.381 & \textbf{0.551} & \textbf{0.486} &  \textbf{0.324} \\
		\textsc{Artif.}\emph{+Emb}  & 0.160 & 0.002 & 0.039 & 0.128 & 0.262 & 0.098\\
		\hline
	\end{tabular}}
\end{table}

The results on the Fairy Tales test data are shown in Table \ref{fairy_table}, still confirming the improvements introduced by constraints in the average case. 
Since ``surprise'' is poorly represented in the labeled portion of the training set (being it not included in ISEAR data), results in this class are pretty low. While artificial labels help in ``surprise'', they sometimes lead to very bad results. This is even more evident when using pre-trained embeddings, where the system constantly overfits the training data. Notice that the F1 scores on the validation splits were very promising when using such embeddings, but, as we mentioned when describing the experimental setting, the system badly generalizes to out-of-sample data that is related-but-not-fully-coherent with the training (validation) sets. 
\begin{table}
	\centering
	\caption{F1 scores on emotion classification (Fairy Tales). Bold: cases in which constraints introduce improvements.}\label{fairy_table}
	\scalebox{0.92}{
	\begin{tabular}{|l|c c c c c c |c|}
		\cline{2-8}
		\multicolumn{1}{c|}{} &  \emph{Anger} & \emph{Disgust} & \emph{Fear} & \emph{Happiness} & \emph{Sadness} & \emph{Surprise} & Macro Avg \\
		\cline{2-8}
		\hline
		\textsc{Plain} & 0.238 & 0.151 & {0.397} & 0.533 & 0.410 & 0.018 &  0.291\\
		\textsc{Constr}  & \textbf{0.288} & \textbf{0.184} & 0.362 & 0.533 & 0.400 & 0.038 &  \textbf{0.301}\\
		\textsc{Artificial} & 0.261 & 0.029 & 0.079 & {0.598} & {0.471} & {0.101} & 0.256 \\
		\hline
		\textsc{Plain}\emph{+Emb}  & 0.365 & {0.137} & {0.451} & {0.546} & 0.365 & 0.037 & 0.317\\
		\textsc{Constr}\emph{+Emb}  & \textbf{0.367} & 0.127 & 0.424 & 0.521 & \textbf{0.476} & \textbf{0.068} & \textbf{0.331}\\
		\textsc{Artif.}\emph{+Emb} & 0.156 & 0.035 & 0.078 & 0.064 & 0.109 & 0.009 & 0.075\\
		\hline
	\end{tabular}}
\end{table} 

In the case of Affect Text test data (Table \ref{affective_table}) constraints still increase the macro F1, but not when using pre-trained embeddings. We observe a less coherent behaviour with respect to the previous test sets, and this is due to the fact that Affective Text is composed of sentences that are significantly shorter than the ones of the other datasets, and they are evocative of multiple emotions in which it is harder to distinguish the most-dominant one.
\begin{table}
	\centering
	\caption{F1 scores on emotion classification (Affective Text). Bold: cases in which constraints introduce improvements.}\label{affective_table}
	\scalebox{0.92}{
	\begin{tabular}{|l|c c c c c c |c|}
		\cline{2-8}
		\multicolumn{1}{c|}{} &  \emph{Anger} & \emph{Disgust} & \emph{Fear} & \emph{Happiness} & \emph{Sadness} & \emph{Surprise} & Macro Avg \\
		\cline{2-8}
		\hline
		 \textsc{Plain}  & 0.162 & 0.100 & {0.282} & 0.514 & 0.289 & 0 &  0.224\\
		 \textsc{Constr}  & \textbf{0.187} & \textbf{0.111} & \textbf{0.282} & 0.493 & 0.295 & 0 &  \textbf{0.228}\\
		 \textsc{Artificial} & 0.182 & 0 & 0.010 & {0.586} & {0.383} & {0.198} & 0.227 \\
		\hline
		\textsc{Plain}\emph{+Emb}   & {0.153} & 0.113 & {0.369} & 0.571 & 0.396 & {0.054} &  {0.276}\\
		\textsc{Constr}\emph{+Emb}   & 0.022 & \textbf{0.117} & 0.324 & \textbf{0.577} & \textbf{0.447} & 0 &  0.248\\
		\textsc{Artif.}\emph{+Emb}  & 0.126 & 0.047 & 0 & 0.059 & 0.093 & 0.045 & 0.062\\
		\hline 
	\end{tabular}}
\end{table}

\begin{figure}[!ht]
\centering
	\includegraphics[width=0.45\textwidth]{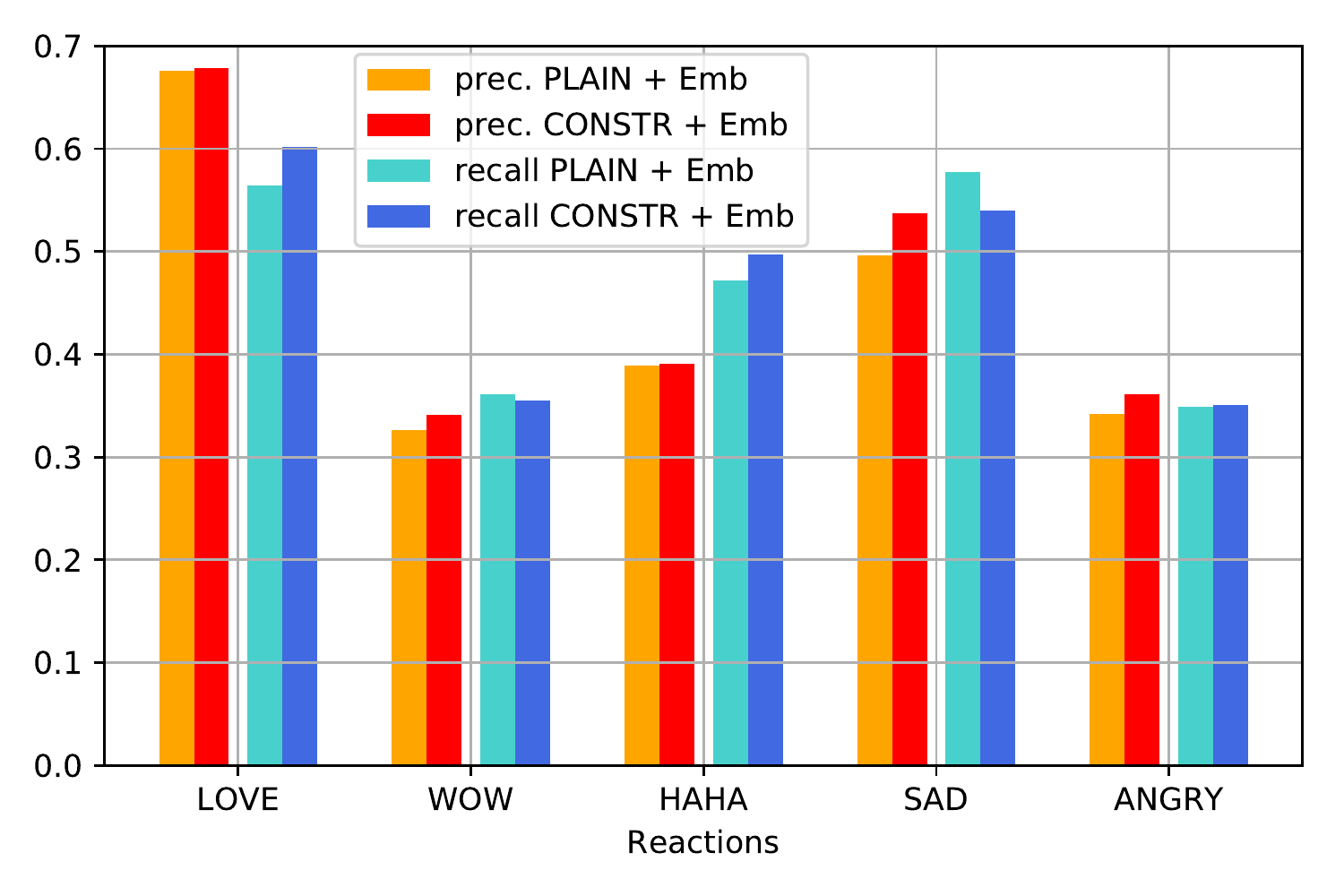}
	\includegraphics[width=0.45\textwidth]{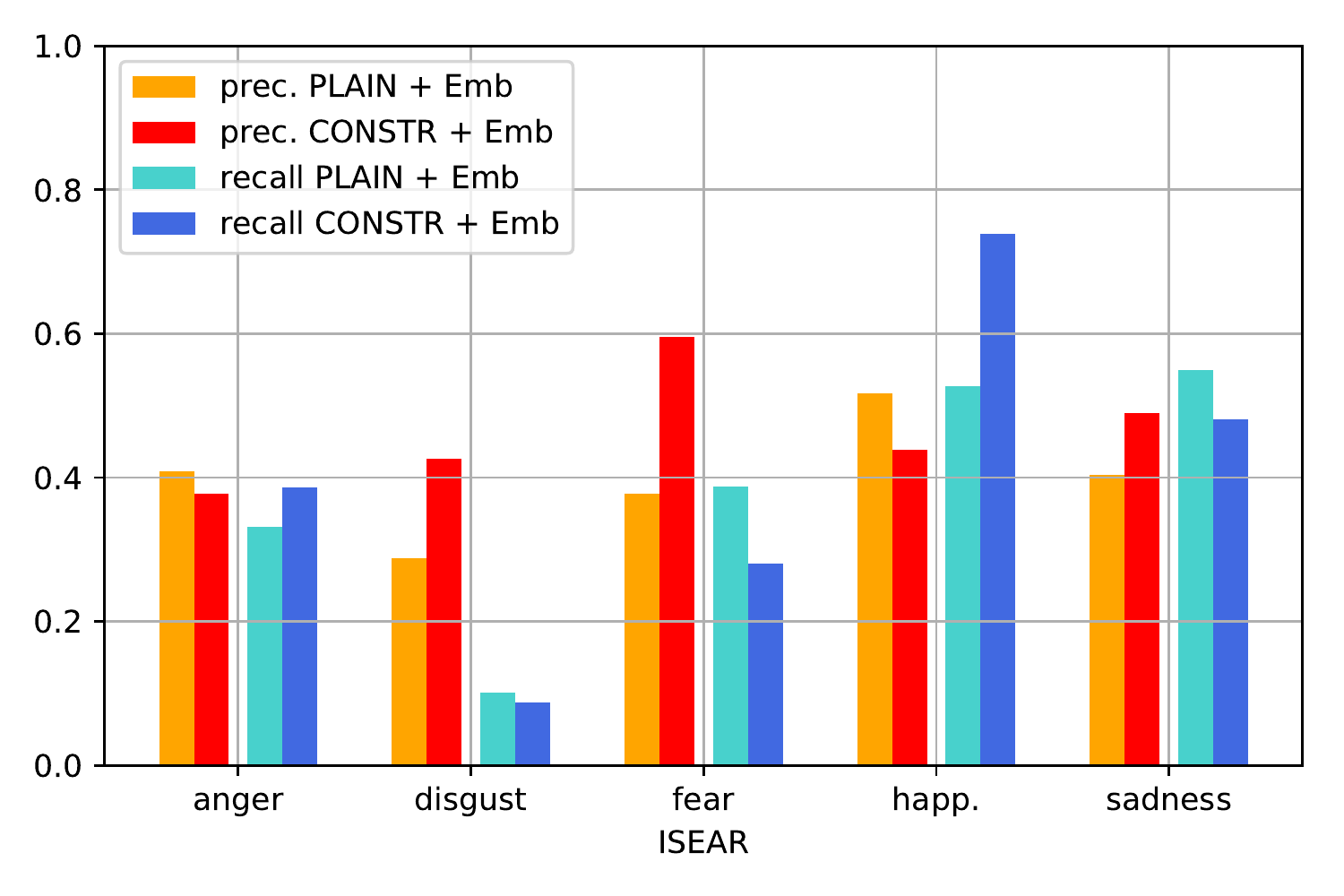}\\
	\includegraphics[width=0.45\textwidth]{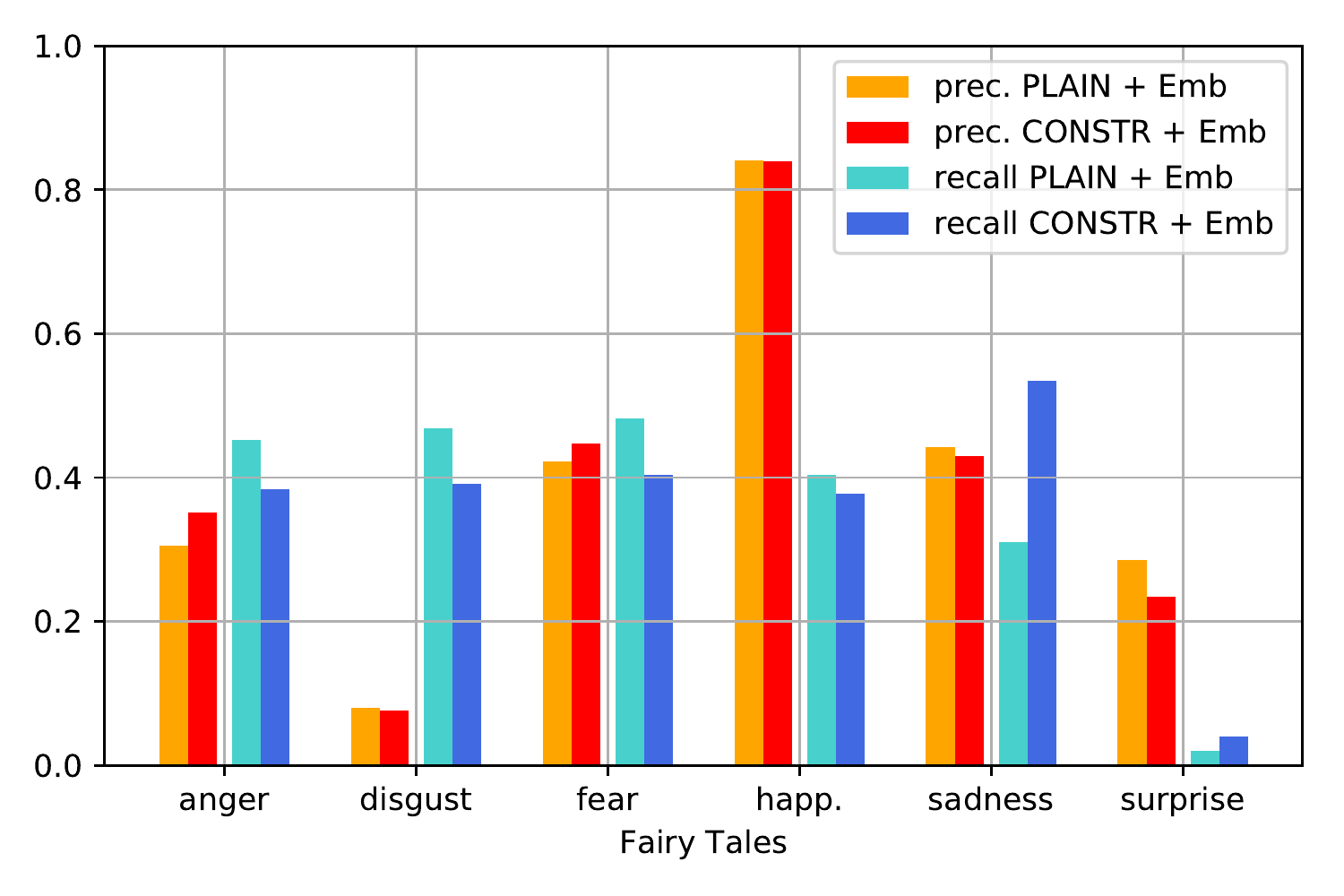}
	\includegraphics[width=0.45\textwidth]{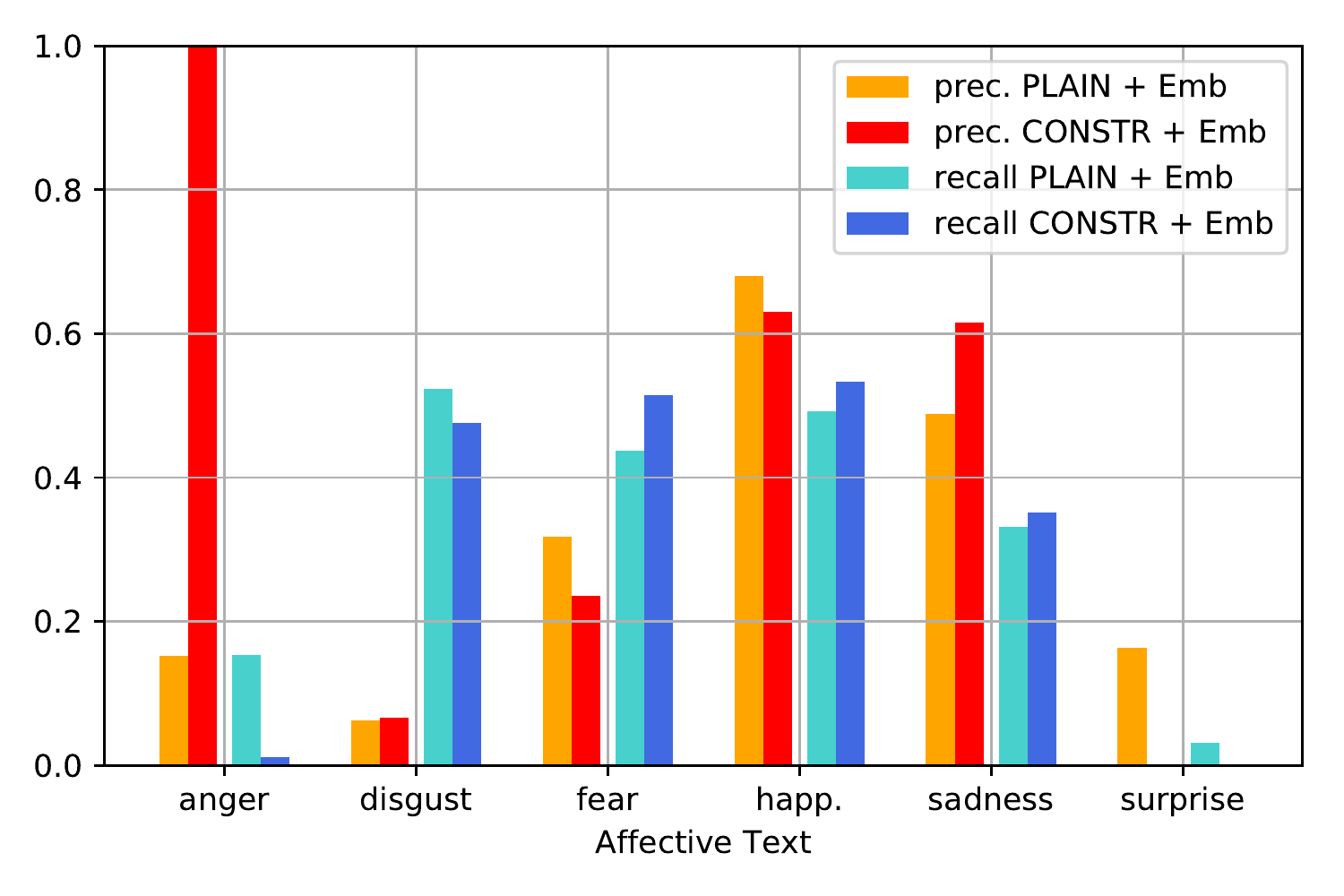}
	\caption{Precision and recall associated to the results of Table~\ref{reaction_table},\ref{isear_table},\ref{fairy_table},\ref{affective_table} (left-to-right, top-to-bottom), comparing \textsc{Plain}\emph{+Emb} with \textsc{Constr}\emph{+Emb}.} \label{prec_rec}
\end{figure}

In Fig. \ref{prec_rec} we report precision and recall (averaged on the test splits, when needed) associated to the results of Table~\ref{reaction_table},\ref{isear_table},\ref{fairy_table},\ref{affective_table}. When predicting reactions and using constraints, we observe improvements in \textit{both} precision and recall in the case of 3 out of 5 classes. When predicting emotions, improvements are usually either in terms of precisions \textit{or} in terms of recall (we count a similar number of cases in which precision is improved and cases in which recall is improved).

Comparing our experimental analysis with existing literature that is about emotion detection is not straightforward.
Existing approaches make use of lexical resources or focus on settings that are pretty different from the one we selected (they test on splits that are taken from the same emotional dataset, thus providing better results \cite{wang2015detecting,mohammad2012portable}). However, we found that, in some cases, our model is competitive with popular algorithms. Table \ref{state_of_art} reports the F1 scores of existing models, emphasizing the cases in which our \textsc{Constr}\emph{+Emb} outperforms them.
In Affective Text, we compared with the WN-AFFECT system (based on WordNet Affect), and a model based on LSA to compute representations of emotion words \cite{strapparava2008learning} (even if they considered a multi-label learning problem). On the same data, as well as in ISEAR, we also considered the CNMF model from \cite{kim2010evaluation}, based on non-negative matrix factorization, that was evaluated on a subset of the emotions we considered in this paper. Finally, we compared with (what we refer to as) the Wikipedia model from \cite{agrawal2012unsupervised}, that was trained on texts taken from Wikipedia and tested on the ISEAR data (and other datasets).\footnote{We did not consider Fairy Tales since existing approaches usually merge ``anger'' and  ``disgust'', and also because the sentence truncation strongly affected this dataset.} 
\begin{table}[!ht]
	\centering
	\caption{Results of existing approaches. We indicate with $*$ those cases in which our model (\textsc{Constr}\emph{+Emb}) outperforms the result reported in this table.}\label{state_of_art}
	\scalebox{0.92}{
	\begin{tabular}{|l|c c c c c c |c|}
		\cline{2-8}
		\multicolumn{1}{c|}{}&  \emph{Anger} &\emph{ Disgust} & \emph{Fear} & \emph{Happiness} & \emph{Sadness} & \emph{Surprise}  & Macro Avg \\
		\cline{2-8}
		\multicolumn{1}{c|}{} & \multicolumn{6}{c|}{ISEAR} & $\ $\\
		\hline
		\textsc{CNMF} \cite{kim2010evaluation} & 0.579 & - & 0.056* & 0.010* & 0.017* & - & -\\
		\textsc{Wikipedia} \cite{agrawal2012unsupervised} &  0.413 & 0.430 & 0.517 & 0.514* & 0.396* & - & 0.454 \\
		\hline
		\multicolumn{1}{c|}{} & \multicolumn{6}{c|}{Affective Text} & $\ $\\
		\hline
		\textsc{WN-AFFECT} \cite{strapparava2008learning} &  0.061 & - & 0.033* & 0.011* & 0.066* & 0.069 & 0.040*\\
		\textsc{LSA} \cite{strapparava2008learning}  &  0.112 & 0.039* & 0.219* & 0.308* & 0.206* & 0.141 & 0.176*\\
		\textsc{CNMF} \cite{kim2010evaluation} &  0.278 & - & 0.618 & 0.648 & 0.475 & - & -\\
		\hline
	\end{tabular}}
\end{table}

\section{Conclusions}
\label{sec:conclusions}

In this paper we proposed to jointly learn the tasks of emotion classification and prediction of Facebook reactions, when processing raw text. While such tasks share several analogies, mapping emotion classes to Facebook reactions (and vice-versa) can easily become ambiguous. Our system exploits First Order-Logic formulas to model the task relationships, and it learns from such formulas, also exploiting large collections of unlabeled training data. The provided experimental analysis has shown that bridging these two tasks by means of FOL-based constraints leads to improvements in the prediction quality that clearly goes beyond more naive approaches in which artificial labels are generated in the data preprocessing stage. Our future work will focus on the introduction of lexical resources in our system.

\subsection*{Acknowledgements}
This project has received funding from the European Union's Horizon 2020 research and innovation program under grant agreement No 825619.

\bibliographystyle{splncs04}
\bibliography{icann2019}

\end{document}